\documentclass[times, 10pt,twocolumn]{article}

\usepackage[utf8]{inputenc}
\usepackage{cvpr}
\usepackage[in]{fullpage}
\usepackage{multicol}
\usepackage{graphicx}
\usepackage{capt-of}
\usepackage{placeins}
\usepackage{subcaption}
\usepackage{filecontents}
\usepackage[backend=biber]{biblatex}
\usepackage{graphicx}
\usepackage{bbm}
\usepackage{amsmath}
\usepackage{mathrsfs}
\usepackage[linesnumbered,ruled]{algorithm2e}
\usepackage{bbm}
\usepackage{booktabs}
\usepackage{graphicx}

\begin{document}
\title{Adversarial Training for Face Recognition Systems using Contrastive Adversarial Learning and Triplet Loss Fine-tuning}
\author{
    Nazmul Karim\\
    \texttt{UCF}\\
    \small{nazmul.karim18@knights.ucf.edu}
    \and
    Umar Khalid\\
    \texttt{UCF}\\
    \small{umarkhalid@knights.ucf.edu}
    \and

    Nick Meeker\\
    \texttt{UCF}\\
    \small{nickmeeker@knights.ucf.edu}
    \and
    Sarinda Samarasinghe\\
    \texttt{UCF}\\
    \small{sarinda@knights.ucf.edu}
    
}

\maketitle

\begin{abstract}
Though much work has been done in the domain of improving the adversarial robustness of facial recognition systems, a surprisingly small percentage of it has focused on self-supervised approaches. In this work, we present an approach that combines Adversarial Pre-Training with Triplet Loss Adversarial Fine-Tuning. We compare our methods with the pre-trained ResNet50 model that forms the backbone of FaceNet, finetuned on our CelebA dataset. Through comparing adversarial robustness achieved with no adversarial training, triplet loss adversarial training, and our contrastive pre-training combined with triplet loss adversarial fine-tuning, we find that our method achieves comparable results with far fewer epochs required during fine-tuning. This seems promising, as increasing the training time for fine-tuning should yield even better results. In addition to this, a modified semi-supervised experiment was conducted, which demonstrated the improvement of contrastive adversarial training with the introduction of small amounts of labels.
\end{abstract}

\section{Introduction}

It is difficult to overstate the importance of facial recognition technologies in the modern world. The most recent survey of state of the art applications employing some sort of facial recognition technology includes systems ranging from video surveillance, criminal identification, building access control, and, increasingly, autonomous vehicle settings, to name just a few \cite{FaceRecogSurvey}. 

With the advancement of deep learning (DL) models in fields like computer vision and natural language processing (NLP), there have been a growing concern about the safety of these models against attacks. One such attack is adversarial attack where one can make imperceptible changes to the input such that the DL model misclassify the modified input. The modified input, also known as adversarial examples, can be generated either accessing the parameters of the neural network or in a black box manner. Efforts have been made to generate such adversarial examples in different ways to fool different types of neural networks. Face recognition models are no exception, as it has been found that deep learning based face recognition models are also vulnerable to adversarial attacks. Very few works have been proposed to increase the robustness of such models. Previously proposed defense strategies mostly require long training on labeled data which is computationally expensive and lacks robustness against strong attacks.

\begin{figure*}[t]
    \centering
    \includegraphics[width=1\linewidth]{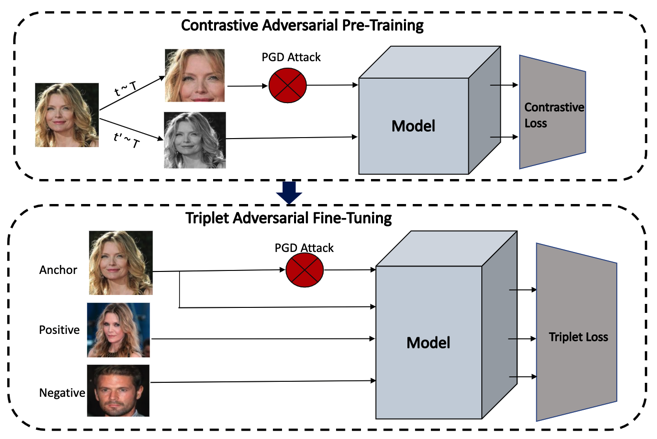}
    \caption{Our proposed approach for improving the robustness of a face recognition model. A contrastive learning based adversarial pre-training is designed for learning robust feature without labels. After that, we use supervised fine-tuning using a triplet loss. Both pre-training and fine-tuning stages employ PGD attack based on contrative and triplet losses respectively.}
    \label{fig:main}
\end{figure*}

In this work, we aim to explore the applications of one such defense strategy -- adversarially robust contrastive learning -- to the field of facial recognition by applying state of the art self-supervised defense techniques to state of the art facial recognition systems. Specifically, we propose a technique for applying self-supervised adversarial pre-training followed by triplet-loss based fine-tuning. In figure \ref{fig:main}, we show our framework that summarizes our approach. By employing two different losses for our Projected Gradient Descent (PGD) attacks, we provide a unique perspective of adversarially training a face recognition model. Learning features that are robust against adversarial attacks is the goal of this work. Adversarial contrastive learning based pre-training offers to capture invariant and useful representations from faces that help us obtain better generalizations for clean images and higher robust accuracy for adversarial images. Our experimental results demonstrate superior performance using both contrastive adversarial pre-training and triplet loss fine-tuning over standard adversarial training. To our knowledge, ours is the first work to propose a pre-training based adversarial robustness approach for face recognition system.

\FloatBarrier

We begin with a presentation of relevant works in the literature before introducing the core concepts of our methodology. We then describe our experimental setup, and offer a presentation of our results.

\section{Related Works}

\subsection{Adversarial Attack Strategies}

Since 2014, when the observation was made that applying small perturbations to inputs can cause dramatic shifts in model outputs\cite{IntriguingPropOfNN}, the field of adversarial machine learning has been elevated to the forefront of computer vision research, with numerous techniques for generating so-called adversarial attacks being published each year. One such technique, FGSM\cite{ExplainHarnessAdv}, is based upon the model gradients, and serves as the foundation for other techniques such as PGD\cite{TowardsDeepLearning}; this class of adversarial example generation technique generally applies per-pixel perturbations budgeted according to some other techniques, such as DeepFool\cite{DeepFool}and C\&W \cite{EvalRobustnessNN}, similarly apply pixel-level changes, often while simultaneously reducing the overall amount of applied perturbation and still maintaining an impressive rate of misclassification. Though many of these pixel-level perturbation techniques are generalizable from the task of image classification to broader applications, there has been some research into adapting these techniques specifically to suit the domain of facial recognition \cite{AdvOnFaceDetec}.

On the other hand, and particularly in the field of facial recognition, another popular method of generating adversarial attacks involves applying perturbations not at the level of the individual pixel, but instead at the level of salient features within the image (that is, the mouth, eyes, etc.). AdvFaces \cite{AdvFaces}, for example, is a genetic algorithm-based approach which generates so-called “perturbation masks” to be applied to face images, resulting in changes in facial features that range from imperceptible to visually apparent; crucially, these perturbations can be generated in either an untargeted (obfuscating) or targeted (impersonating) manner. SemanticAdv \cite{SemanticAdv} generates similar perturbations by employing a semantic segmentation strategy and applying transformations to salient semantic regions (hairlines, for example). In a similar vein, GFLM \cite{FastGeomPerturbAdv} applies geometric scaling to salient image regions, with horizontal and vertical transformation hyperparameters controlling the level of applied perturbation. 

Crucially, all of the aforementioned techniques are supervised adversarial techniques; that is, they require labeled datasets to successfully generate perturbations. The burgeoning field of self-supervised adversarial generation is a relatively recent and active area of research \cite{AdvSelfSupContr}. 

\subsection{Adversarial Defense Strategies}

Arguably the most thoroughly and successfully researched adversarial defense strategy is that which involves adversarial training, or retraining the attacked model against adversarial examples in order to improve its adversarial robustness (all of the adversarial training citations there are a bunch). This increase in robustness is generally accompanied by a notable decrease in performance against benign inputs, leading to a well-established trade-off between robustness and accuracy \cite{RobustnessVAcc}\cite{RobustnessOddsAcc}. Furthermore, adversarial training itself raises concerns with respect to its scalability as a practice, particularly regarding the transferability of adversarial robustness; models trained against one set of adversarial examples may not be resistant to examples generated with a different technique, resulting in the need for repeated retraining \cite{AdvMLAtScale}. 

Though adversarial training is certainly the most popular avenue of exploration in improving adversarial robustness, other techniques have been suggested. One such technique involves applying random resizing and padding during input evaluation, and in doing so achieves an impressive degree of robustness against standard adversarial attack techniques \cite{RandomResizing}, though this method suffers from a weakness to translation-invariant adversarial attack strategies \cite{EvadingDefTransInvAtk}. In the domain of facial recognition specifically, another common avenue of defense involves applying a pre-processing purification technique to the adversarial input, aiming to systematically remove perturbations, though these techniques can often be subject to exceptionally poor classification rates against benign images due to their tendency to “purify” benign images (removing perturbations where there are none) \cite{FaceGuard}.

\section{Methods}

\subsection{Contrastive Loss}

We first introduce the concept of contrastive loss, which serves as the foundation for self-supervised contrastive adversarial learning. Let sim($u$, $v$) be a measure of the similarity (e.g., cosine similarity) between $u$ and $v$. Consider a positive pair of images, or two pairs belonging to the same labeled class, ($i$, $j$). The contrastive loss for this pair, then, can be computed as
\begin{equation}
    \ell_{i,j} = -\text{log}\frac{\text{exp}(\text{sim}(z_i,z_j)/\tau)}{\sum_{k=1}^{2N}\mathbbm{1}_{[k\neq{i}]}\text{exp}(\text{sim}(z_i, z_j)/\tau)}\nonumber
\end{equation}

where $\mathbbm{1}_{[k\neq{i}]} \in \{0, 1\}$  represents an indicator evaluating to $1$ iff $k\neq i$, $\tau$ represents temperature, and $z_i$ and $z_j$ represent model output for inputs $i$ and $j$ \cite{SimCLR}.

\subsection{Self-Supervised Contrastive Learning}

Self-supervised contrastive learning aims to learn general features on a dataset without labels. This is done by focusing on the similarity between images within a class and increasing the distance between classes. For each image in a dataset, augmented, positive versions are generated through cropping, resizing, recoloring, etc. These augmented versions of the image will ideally have the same classification as the original image. By using a contrastive loss function, the difference between the model’s encodings of an image and its augmented versions is minimized. 

In the example of SimCLR, a contrastive learning framework, the method uses random cropping followed by resizing the original image size, randomized color distortions, and random Gaussian blurs, as the three sequential augmentations to generate similar examples. Once the extra images are created, ResNet-50 is used to create the vector representations of the inputs. The projector used is two-layer multi-layer perception (MLP), and the loss function is Normalized Temperature-scaled Cross Entropy Loss (described in the previous section).

\subsection{Instance-wise Attacks}

As opposed to class-wise attacks, instance-wise attacks generate perturbations for single inputs to cause the model to misclassify it as a different sample. This type of attack is used for unlabeled sets of data, which would lack the class distinction for a class-wise attack. These attacks attempt to maximize the self-supervised contrastive loss between the perturbation and the anchor image. For our work, we employ instance-wise adversarial attacks with PGD during adversarial contrastive pre-training, following the methods proposed in \cite{jiang2020robust}.

\subsection{Self-Supervised Contrastive Adversarial Training}

A key aspect of our methodology is the exploration of self-supervised contrastive adversarial training as a pre-training technique applied to facial recognition systems. In this stage, we generate instance-wise adversarial attacks using PGD with contrastive loss, and apply contrastive learning as a vehicle for adversarial training. Specifically, we aim to apply contrastive loss to a clean image, $x$, and its instance-wise adversarial example, $x_{adv}$, for all images in the dataset. 

\begin{algorithm}
    \SetKwInOut{Input}{Input}
    \SetKwInOut{Output}{Output}

    \Input{A set of clean images $X$; Augmentation family $\tau$; Network backbone $f$}
    \Output{Parameters $\theta$ in $f$}
    \For{$x$ in $X$}
        {
            Augment $x$ to be $(\Tilde{x})$ with augmentation sampled from $\tau$
            
            Generate the corresponding adversarial example $(\Tilde{x} + \delta)$ with
            \begin{equation}
                \delta = \underset{||\delta||_\infty \leq \epsilon}{\text{arg max}} \ell_{CL}(f(\Tilde{x} + \delta); \theta, \theta_{adv}) \nonumber
            \end{equation}
            
            $\ell = \ell_{CL}(f(\Tilde{x},\Tilde{x} + \delta); \theta, \theta_{adv})$
            
            Update parameters $(\theta, \theta_{adv})$ to minimize $\ell$
        }
    \caption{Algorithm for Contrastive Adversarial Pre-training }
\end{algorithm}

Algorithm 1 describes the proposed method in detail. The technique is fundamentally similar to the A2S technique proposed in \cite{jiang2020robust}, but with the crucial difference that it is employed here as pre-training technique.

\subsection{Triplet Loss}

Augmented images that are generated in self-supervised contrastive learning are called positive images, as they should be associated with the base, or anchor image. Any images belonging to or formed from a differing class are negative images. To calculate triplet loss on a model, multiple triplets are formed and compared after they are fed through the model. Each triplet consists of an anchor image, a positive image, and a negative image. Intuitively, the encodings of the anchor and positive images would ideally be much more similar than the negative image is to either. The triplet loss function returns the difference of the distances between the anchor and positive images and the anchor and negative images, or zero if the difference is negative.

\FloatBarrier
\[L(a,p,n) = max(0, D(a,p) - D(a,n) + margin)\]
\captionof{figure}{Triplet Loss Function}
\FloatBarrier

\subsection{Triplet Loss Adversarial Fine-tuning}

In this work, we apply triplet loss-based adversarial fine-tuning following self-supervised adversarial pre-training. Triplet loss-based adversarial fine-tuning is a supervised technique, and involves gathering an anchor image, positive image, and negative image for each input in the dataset. In order to incorporate adversarial training into triplet loss-based learning, we propose a new technique for generating anchor images for each input in the dataset. 

First, for each image $x$, a PGD adversarial example, $x_{adv}$, is generated. In this stage of adversarial training, PGD adversarial examples are generated using triplet loss. In order to form the anchor for triplet learning, $a$, $x_{adv}$ is concatenated to $x$, with the resulting output serving as the anchor. To our knowledge, ours is the first work to investigate generating adversarial examples against facial recognition systems by leveraging triplet loss in this manner. 

Algorithm 2 describes our technique in detail.

\FloatBarrier
\begin{algorithm}[htbp]
    \SetKwInOut{Input}{Input}
    \SetKwInOut{Output}{Output}

    \Input{A set of clean images $X$; A set of positive images $P$; A set of negative images $N$; Network backbone $f$}
    \Output{Parameters $\theta$ in $f$}
    \For{$x$ in $X$}
        {
            Select positive image $p$ from $P$ and negative image $n$ from $N$
            
            Generate the corresponding adversarial example $(\Tilde{x} + \delta)$ with
            \begin{equation}
                \delta = \underset{||\delta |_\infty \leq \epsilon}{\text{arg max}} \ell_{TL}(f(\Tilde{x} + \delta, p, n); \theta, \theta_{adv}) \nonumber
            \end{equation}
            
            Generate the anchor image $a = x^\frown (x + \delta)$
            
            $\ell = \ell_{TL}(f(a,p,n); \theta, \theta_{adv})$
            
            Update parameters $(\theta, \theta_{adv})$ to minimize $\ell$
        }
    \caption{Algorithm for Triplet Loss Adversarial Fine-tuning }
\end{algorithm}
\FloatBarrier
 
\section{Experiments}

\subsection{Datasets}

CelebFaces Attributes Dataset (CelebA) consists of over 200,000 images of celebrities, each with name labels and binary attributes for the appearance of various body part types (mustache, wavy hair, etc.) and clothing (glasses, hat, etc.). For initial training purposes, the whole dataset is used, but for later experiments with contrastive learning, the dataset is filtered to include only classes (individual people) for whom multiple images are available. This is because triplet loss, discussed in a later section, requires at least two images for each class. After this filtering process, 9,136 remain available, with 8,156 designated for training and 980 for validation.

\subsection{Triplet Loss Adversarial Training}

To establish a baseline by which to compare our proposed methodology (adversarial pre-training with contrastive loss followed by triplet loss-based adversarial fine-tuning), we first examine the effect of triplet loss adversarial training alone on model robustness. In order to generate adversarial examples, an $\ell_{\infty}$-norm PGD attack with triplet loss is used, with $\epsilon = 8/255$, $\alpha = 2/255$, and maximum iterations set to 7. The model, already pre-trained on clean images from our dataset, is then adversarially trained using the triplet loss technique described in Algorithm 2 for 250 epochs. A batch size of 128 is used for training.

\subsection{Self-Supervised Contrastive Adversarial Pre-training with Triplet Loss Fine-tuning}

In this stage, we examine the effectiveness of applying self-supervised contrastive adversarial pre-training, and then fine-tuning afterwards in a supervised manner using triplet loss-based adversarial training. For contrastive adversarial pre-training, instance-wise adversarial examples are generated using an $\ell_{\infty}$-norm PGD with contrastive (as opposed to triplet) loss, maintaining all the same hyperparameters described above. The model is then trained for 500 epochs in a self-supervised contrastive manner before being fine-tuned for 50 epochs using triplet loss-based adversarial training. This is analogous to performing 500 epochs of Algorithm 1 followed by 50 epochs of Algorithm 2, both described above. Again, a batch size of 128 is used, both for the contrastive adversarial training step and for the triplet loss adversarial fine-tuning step. 

\subsection{Semi-Supervised Contrastive Adversarial Pre-training with Triplet Loss Fine-tuning}

As a final point, \cite{jiang2020robust} observes that introducing even small amounts of labels (1\%, 10\%, etc.) to an otherwise unlabeled dataset can dramatically improve the results of contrastive adversarial training. Following this observation, we examine the effect of performing semi-supervised adversarial contrastive training followed by triplet loss adversarial fine-tuning. To achieve this, we recreate the experiment setup from the prior section, but reintroduce labels during the contrastive adversarial learning stage. All other hyperparameters remain consistent with those described in the previous section.

\section{Results}

For the following subsections we compare Standard Training, Triplet Adversarial Training, and Contrastive Pre-Training + Triplet Adversarial Fine-Tuning. Standard training is simply the base pretrained Resnet50 model from FaceNet that was fine-tuned to the CelebA dataset.

For our accuracy metrics we have Standard Accuracy (SA), Robust Accuracy (RA), and Standard \& Robust Accuracy (R\&A). Standard accuracy is accuracy on a clean dataset with no perturbations, Robust Accuracy is accuracy on the perturbed dataset, and Standard \& Robust Accuracy is the evaluation over the two combined. 

An interesting result that is apparent in the following tables is that for the adversarial training techniques, the robust accuracy is greater than the standard accuracy. At face-value this is not very intuitive as the introduction of perturbations should impede performance if they are meaningful attacks. However, our counter to this is that for an unattacked triplet, the positive and corresponding anchor images are of the same person, but are different pictures. In an attacked triplet, a positive image can be an attacked version of the anchor image. For a model that lacks adversarial training, it would have a hard time comparing the anchor with its attacked counterpart, as it is designed to be as distanced from the anchor as possible. But for a model that has been adversarially trained, it would likely find the perturbed positive image to be even more similar to the anchor than a clean positive image, since the perturbed image is based on the original, instead of being a different image of the same person.

\subsection{Contrastive Loss}

There is no accuracy metric for the contrastive loss for the experiment, but we did see that in training for 500 epochs, the loss began to stabilize around 150 epochs into the training.

\FloatBarrier
\begin{figure}[h!]
\centering
\includegraphics[scale=0.45]{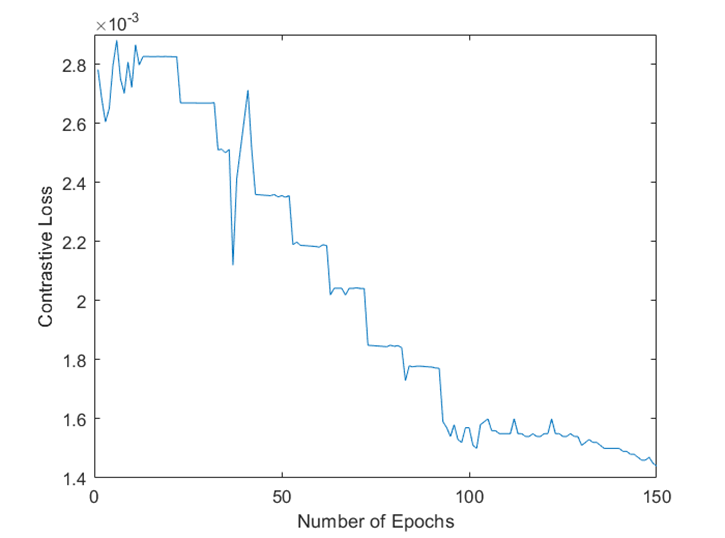}
\caption{Self-Supervised Contrastive Loss }
\end{figure}
\FloatBarrier

\subsection{Self-Supervised Contrastive Adversarial Pre-training with Triplet Loss Fine-tuning}
Both Triplet Adversarial Training and Pre-Training + Fine-Tuning considerably outperformed the Standard model in terms of RA, while slightly lagging behind in terms of SA. This is generally expected when comparing models with and without adversarial training. When comparing our approach with the Triplet Adversarial Training baseline, we found that our results were very similar to the baseline, despite training for much less through fine-tuning. We firmly believe that through further fine-tuning, our model can easily outperform the baseline in this scenario, but we were unable to in the interest of time.

\FloatBarrier
\begin{figure}[h!]
\centering
\includegraphics[scale=0.60]{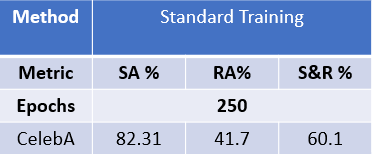}
\includegraphics[scale=0.60]{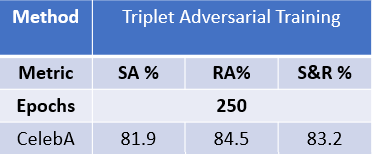}
\includegraphics[scale=0.60]{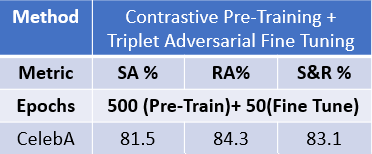}
\caption{Triplet Adversarial Training}
\end{figure}
\FloatBarrier

\subsection{Semi-Supervised Contrastive Adversarial Pre-training with Triplet Loss Fine-tuning}
With the introduction of a subset of labels, our model completely surpasses both FaceNet and the Triplet baseline, achieving roughly 30\% higher accuracy across all datasets.

\FloatBarrier
\begin{figure}[h!]
\centering
\includegraphics[scale=0.60]{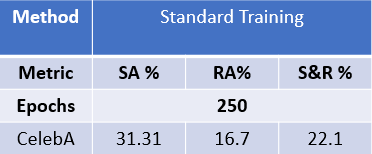}
\includegraphics[scale=0.60]{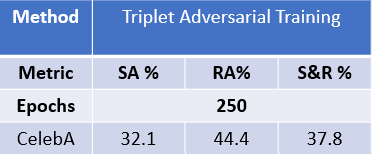}
\includegraphics[scale=0.60]{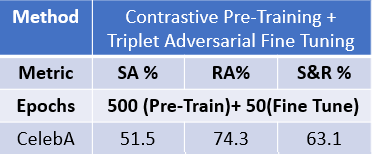}
\caption{Semi-Supervised Contrastive Adversarial Pre-training with Triplet Loss Fine-tuning}
\end{figure}
\FloatBarrier

\section{Conclusion}

In this work, we have proposed a methodology for improving the adversarial robustness of facial recognition systems by applying adversarial contrastive pre-training and triplet loss-based adversarial fine-tuning. Crucially, we observe that, by applying unsupervised adversarial pre-training, we are able to dramatically reduce the number of epochs required for triplet-loss based adversarial training in order to produce robust results. We also demonstrate that results can be improved further by reintroducing labels during the adversarial contrastive pre-training phase. On the whole, our contrastive adversarial training approach offers a label-efficient technique for improving adversarial robustness of facial recognition systems, and for reducing the number of adversarial training epochs required during fine-tuning stages.

As avenues of future work, we would first consider investigating the defense of a wider variety of models; in this work, we applied our technique only to ResNet50 pre-trained and fine-tuned for facial recognition. There are certainly more, larger facial recognition architectures in the literature, and it would be worth investigating if our results are equally as effective when applied to them. Additionally, we would consider investigating against a more diverse array of stronger adversarial attack strategies. For example, we would consider testing against the geometric-based adversarial attacks discussed in the literature review. Finally, we would investigate the transferability of our robustness results, specifically investigating if models that are adversarially trained using our approach remain more resistant to adversarial attacks to which they were not exposed during training. 

\section{Source Code}

All source code for this project has been made available at https://github.com/UmarKhalidcs/ACL-for-Face-Recognition.

\printbibliography

\end{document}